\documentclass{article}
\usepackage{ijcai21}
\usepackage{times}
\usepackage{soul}
\usepackage{url}
\usepackage[hidelinks]{hyperref}
\usepackage[utf8]{inputenc}
\usepackage[small]{caption}
\usepackage{graphicx}
\usepackage{amsmath}
\usepackage{amsthm}
\usepackage{booktabs}
\usepackage{algorithm}
\usepackage{algorithmic}
\usepackage{epsfig}
\usepackage{amsmath,bm}
\usepackage{amssymb}
\usepackage{hyperref}
\usepackage{amsfonts}
\usepackage{nicefrac}
\usepackage{microtype}
\usepackage{pgfplots}
\usepackage{pgfplotstable}
\usepackage{filecontents,pgfplots}
\usepackage{tikz}
\usepackage{mathrsfs}
\usepackage{multirow}
\usepackage{comment}
\usepackage{rotating}
\usepackage{mwe}

\newcommand{\DD}{\mathcal{D}}
\newcommand{\UU}{\mathcal{U}}
\newcommand{\NN}{\mathcal{N}}
\newcommand{\ie}{\textit{i.e.}}
\newcommand{\eg}{\textit{e.g.}}

\hypersetup{colorlinks=true, linkcolor=red}

\usetikzlibrary{arrows,calc,shapes,snakes,positioning,matrix,arrows,decorations.pathmorphing,decorations.text}

\title{Image/Video Deep Anomaly Detection: A Survey}

\author{
Bahram Mohammadi$^1$, Mahmood Fathy$^2$
\And
Mohammad Sabokrou$^3$
\affiliations
$^1$Sharif University of Technology \\
$^2$Iran University of Sciences and Technology (IUST) \\
$^3$Institute for Research in Fundamental Sciences (IPM)
\emails
bmohammadi@alum.sharif.edu, mahfathy@iust.ac.ir,
sabokro@ipm.ir
}

\begin{document}

\maketitle

\begin{abstract}
  The considerable significance of Anomaly Detection (AD) problem has recently drawn the attention of many researchers. Consequently, the number of proposed methods in this research field has been increased steadily. AD strongly correlates with the important computer vision and image processing tasks such as image/video anomaly, irregularity and sudden event detection. More recently, Deep Neural Networks (DNNs) offer a high performance set of solutions, but at the  expense of a heavy computational cost. However, there is a noticeable gap between the previously proposed methods and an applicable real-word approach. Regarding the raised concerns about AD as an ongoing challenging problem, notably in images and videos, the time has come to argue over the pitfalls and prospects of methods have attempted to deal with visual AD tasks. Hereupon, in this survey we intend to conduct an in-depth investigation into the images/videos deep learning based AD methods. We also discuss current challenges and future research directions thoroughly.
\end{abstract}

\begin{figure}[t]
    \center
	\includegraphics[width=0.95\linewidth]{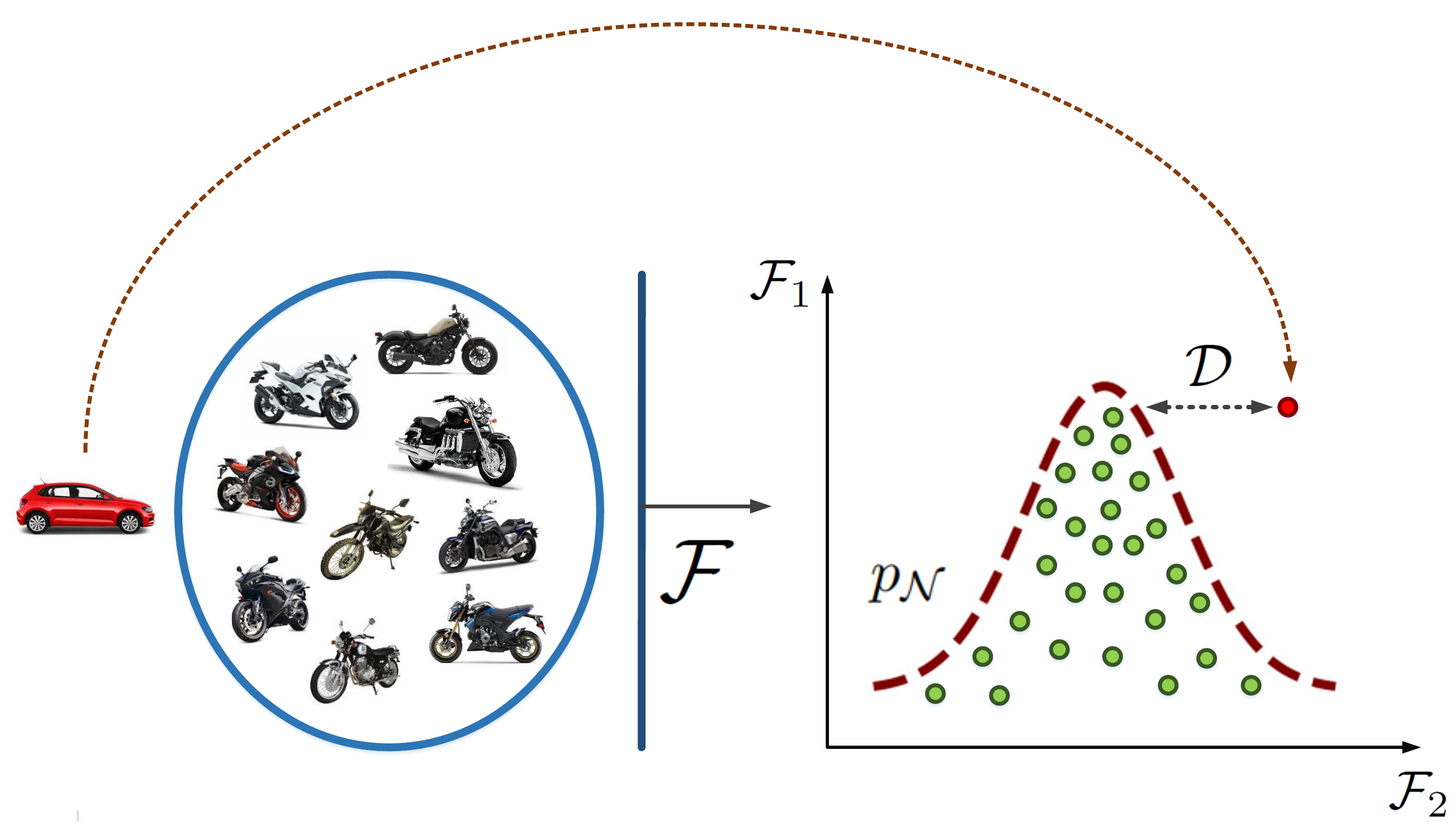}
	\caption{The general concept of AD is depicted in this figure. Here, motorcycles are considered as normal instances while the car is anomaly. $\mathcal{F}$ demonstrates a representation of the given data for analysis. For simplicity, samples are shown in two dimensions using $\mathcal{F}_1$ and $\mathcal{F}_2$ feature vectors. As is clear, motorcycles, which are denoted by green dots, follow the distribution of target class (normality), \ie, $p_\mathcal{N}$. Therefore, an out-of-distribution instance (car in our case), which is represented by a red dot, has a deviation from the normal data calculated by a specific detection measures, \ie, $\mathcal{D}$.}
	\label{fig:ad}
\end{figure}

\section{Introduction}
Anomaly Detection (AD) is the task of detecting samples and events which rarely appear or even do not exist in the available training data. Indeed, AD is the process of looking for the unseen concepts. Generally, in the context of AD, there are a huge number of data instances following the target class distribution, \ie, normal data. On the other hand, samples belonging to the out-of-distribution class (outliers) are not present or scarcely accessible but at the expense of high computational cost. In summary, abnormalities can derive from any unknown distribution which leads to a very complicated learning process. Hence, instead of learning irregularities, researchers have proposed to distill the shared concept among all of the normal data as one (several) reference model(s) \cite{bertini2012multi,sabokrou2015real}. In the stage of testing, the deviation of an instance from such model(s) shows whether it is an anomaly or not. Figure \ref{fig:ad} shows an overall sketch of the AD concept. 

With respect to the type of data, an AD task may encounter various difficulties. Common weaknesses that AD algorithms suffer from are, (1) high false positive rate: in the most of AD applications, detecting abnormal events is considered more important and critical than recognizing normal data. For example, in surveillance systems ignoring just one anomalous behaviour, \ie, detecting an anomaly as a normal event, completely compromises the reliability and also the safety of the monitoring system. As a consequence, tolerating a bit more false positive rate is reasonable in order to confidently detect all of the outliers. Nevertheless, high false alarm rate brings unreliability and ineffectiveness, (2) high computational cost: most of the previous works are too complex to quickly and appropriately act in real-world applications and (3) unavailability of a standard dataset for assessment: available datasets are very far from what exist in realistic situations. In fact, in order to comprehensively investigate a proposed solution in this research area, having access to a more real and representative dataset is crucial. The above-mentioned shortcomings confirm that AD tasks face several ongoing challenges which need to be addressed effectively. Furthermore, very recent proposed methods have merely focused on the performance in a simple scenario. Considering the image/video AD methods in different aspects is a key step to improve the current state-of-the-arts.

Inspired by the resounding successes of Deep Neural Networks (DNNs) in different fields of research, a bunch of deep learning based solutions have been presented to deal with AD tasks. Some of them have attained a great performance. Whereas, difficulties in implementation and reproducibility, especially those are based on Generative Adversarial Networks (GANs) \cite{goodfellow2014generative}, along with a high computational overhead are still considered as serious challenges. 

\noindent
\textbf{Scope of Survey.}
Some informative and valuable surveys has been provided thus far. We briefly mention some of more recent works in this area. \cite{chalapathy2019deep} has focused on deep AD for different tasks such as intrusion detection systems, video surveillance, medical and etc. \cite{ruff2019deep} presents a framework for deep AD along with experimental scenarios for the general semi-supervised AD problem. Different deep learning detection techniques for video AD are covered by \cite{suarez2020survey}. To bridge the existing gaps, we present a novel taxonomy for deep learning based image/video AD. We highlight the unsupervised approaches owing to their generalizability, applicability in real-world problems and rising popularity. Having investigated each of categories and state-of-the-art methods specifically, we express the challenging aspects, open problems and effective direction for future works in image/video AD tasks.

\section{Problem Formulation}
In general, there are $\UU$ unlabeled images or video frames, denoted as $X_{\mathcal{N}}$, with the assumption that the majority (not all) of them comply with the distribution of normal data ($p_{\mathcal{N}}$), \ie, $(x \in X_{\mathcal{N}}) \sim p_{\mathcal{N}}$. AD can be considered as the process of realizing whether a test sample like $y$ follows $p_{\mathcal{N}}$ or detected as abnormality. 
\begin{equation}
\begin{split}
    {\text{AD}(\mathcal{F}(y))}
    \begin{cases}
	    \text{Normal} & \mathcal{D}(\mathcal{F}(y),p_{\mathcal{N}})\leq \tau \\
	    \text{Anomaly} & Otherwise
    \end{cases}
\end{split} 
\label{eq:f}
\end{equation}
Where $\DD$ is a metric to compute the distance between given instances and the distribution of normal data. $\mathcal{F}$ is a feature extractor that maps the raw data to a set of discriminative features. According to the number of available normal ($\NN$), anomalous ($\mathcal{A}$) and unlabeled ($\UU$) samples in the training set, proposed methods can be categorized into three types: (1) supervised (2) semi-supervised  and (3) unsupervised (see Section \ref{sec:ssu}). The unsupervised techniques are more effective and applicable in realistic situations.  

Due to the high dimensionality and high diversity of data instances, the explicit learning or fitting a distribution ($p_{\mathcal{N}}$) and utilizing a distance metric ($\DD$) are not straightforward. Learning or selecting a discriminative representation from raw data, $\mathcal{F}$, and fitting a machine learning based approach to learn $p_\NN$ and $\DD$ are necessary steps to introduce an efficient AD solution. Different viewpoints on each of the mentioned steps are investigated by researchers resulting is in a wide range of proposed algorithms (see Sections \ref{sec:sec2} and \ref{sec:sec3}).

\subsection{$p_\NN$: Modeling Normal Data}
As stated previously, the widely-used and well-investigated approaches for AD have considered learning the shared concepts of normal samples as a reference model or a distribution. The first efforts for modeling suggested fitting a predefined distribution, \eg, Gaussian distribution \cite{sabokrou2018deep}. However, regarding the high dimensional instances and consequently the complexity of such data distribution, DNNs, notably GANs, are exploited to implicitly learn the desired distribution \cite{kim2020gan}.

\subsection{$\DD$: Detection Measure}
Regarding the Equation \ref{eq:f}, the measure $\DD$ necessities distinguishing deviated data from the normal reference model, p$_\NN$. Earlier methods used traditional measures like Mahalanobis distance or probability. Recently, proposed solutions implicitly learn both the reference model and detecting measure using DNNs such as encoder-decoder networks and GANs.

\section{Supervision}
\label{sec:ssu}
Basically, AD can be interpreted in different points of view such as availability of data labels, input data nature and training objectives \cite{chalapathy2019deep}. Naturally, AD is the task of training a model to recognize the out-of-distribution data using just unlabeled instances. Nevertheless, in certain circumstances, the availability of labeled samples for both normal and abnormal data is plausible. Accordingly, based on the supervision, \ie, availability of data labels, there are three major categories which are completely explained in the following subsections.

\noindent
\textbf{Supervised ($\NN + \mathcal{A}$).} In some cases and depending on the applications (\eg, fall or accident detection \cite{kong2019learning}), there is a well-defined explanation of abnormality. Thus, gathering $\NN$ normal and $\mathcal{A}$ anomalous samples for training a binary classifier is straightforward. A Convolutional Neural Network (CNN) can be learned on $\NN + \mathcal{A}$ instances in a supervised setting to efficiently make a distinction. Although supervised approaches produce highly accurate results, their outcomes are not sufficiently generalized. The performance of deep supervised classifier is sub-optimal owing to the class imbalance, \ie, the total number of samples belonging to the target class are far more than the whole irregular classes of data \cite{kim2020gan}. Class imbalance ($\mathcal{A} << \NN$) is problematic even in case of having an exact definition of abnormal events. Moreover, remarkable diversity of anomalies disturbs the proper training procedure and practically makes it infeasible. This range of solutions are applicable just to a very limited real-world problems.

\noindent
\textbf{Semi-supervised ($\NN + \mathcal{A} + \UU$).} There are numerous unlabeled data in AD tasks while collecting anomalous instances is a cumbersome and expensive process. This problem arises because abnormalities are very diverse and rarely occur. However, to take advantage of abnormal and normal concepts, some works \cite{ruff2019deep,liu2019margin} propose learning a model on copious unlabeled samples along with a few number of irregular and normal data ($\NN + \mathcal{A} << \UU$). Broadly speaking, almost in all of the AD applications having access to complete range of anomalous and normal events for training is implausible and computationally expensive.  

\noindent
\textbf{Unsupervised ($\UU$).}
Training a model in an unsupervised setting means that the AD task is applied to unlabeled data. In this technique, outliers are detected solely based on intrinsic properties of data instances. The only assumption can be made in this category is that, naturally abnormal events are rarely occurred or appeared in the unlabeled samples just like in realistic situations. Answering to this question determines why we mainly concentrate on unsupervised technique, \emph{"why is unsupervised widely adopted?"} Considering no strict presupposition about the training data leads to generalizability. Since anomalies occur very occasionally in the available data, specifying real abnormalities is a very costly and time-consuming procedure, especially for applications in which there is not a precise definition for anomalous events. Even in case of accessing to anomalies, lack of sufficient data is problematic. Unsupervised methods can be considered as a One-Class Classification (OCC) task. From this perspective, we are able to investigate AD problems in a more general way. 

Some of researchers has argued that any assumption about the label of samples contradicts the unsupervised setting. In fact, this belief is not correct owing to the nature of normal and abnormal data instances. If we haphazardly gather data for various tasks, large amount of data will be normal confidently. Since our central focus is on unsupervised solutions, we thoroughly explain the main deep learning based techniques belonging to this category for image/video AD.

\section{Deep Image/Video Representation}
\label{sec:sec2}
\subsection{Traditional Features}
The first generation of proposed methods for image/video AD was based on either trajectory based features \cite{morris2011trajectory} or low level features such as Histogram Of Gradient (HOG) , Histogram Optical Flow (HOF) and Motion Boundary Histogram (MBH). These approaches suffer from several weaknesses such as high computational cost and low performance, in spite of their power in classification. Furthermore, they are not adequately discriminative and usually suffer form high false positive rate. Hand-crafted features are also ineffective to properly deal with occlusion.

It is worth mentioning that, both spatial and temporal characteristics of a video play the vital role for accomplishing AD tasks. There are widely-used techniques such as Recurrent Neural Network (RNN), Long Short Term Memory (LSTM) and 3D-CNN to get involve temporal features as a part of deep learning based method for video AD. With respect to the limited space, we only concentrate on the key ideas for AD and thus details of DNN architectures such as number of layers, kernels and how the temporal features are represented are not discussed.

\subsection{Deep Features}
Impressive achievements of deep representation learning on a wide range of computer vision tasks encourages researchers to take advantage of learned features instead of using the traditional hand-crafted ones. We group this technique into two approaches: feature learning and pre-trained networks.

\noindent
\textbf{Feature Learning.}
Early presented methods have utilized auto-encoders as a popular tool for providing a discriminative representation \cite{xu2017detecting,sabokrou2015real,sabokrou2018deep}. In fact, they only modified the traditional solutions by using the learned feature set instead of hand-crafted features. For instance, \cite{xu2018unsupervised} proposed the usage of stacked de-noising auto-encoders to automatically learn both appearance and motion feature sets which have further been exploited as an input for multiple one-class SVM models. In this regard, \cite{bertini2012multi,sabokrou2018deep,sabokrou2017fast} explain that the processing an image or a video frame entirely and in just one phase results in a high computational cost. To tackle this problem, the first approaches suggested dividing the images to a set of patches to perform as a patch based algorithm. Learning features of each patch of a frame was done by auto-encoders in earlier solutions. In this case, the encoder is able to represent all of the patches while the obtained feature set is more discriminative. Thanks to the generality of deep features, this way of approaching the problem has achieved better results than the traditional methods. Processing the images/frames in a patch based manner has raised two critical difficulties: (1) dividing the images/frames into a set of patches and representing them one by one is considered as overhead and (2) it is a ad hoc setting whereas the community is very interested in end-to-end deep learning structures.

\noindent
\textbf{Pre-trained Networks.}
In the context of AD, transfer learning is capable of efficiently follow the idea of distilling more domain knowledge into a model, \eg, through using and possibly fine-tuning pre-trained networks. \cite{sabokrou2017deep} realized that the first layers of a pre-trained DNN like AlexNet \cite{krizhevsky2012imagenet} is sufficiently informative for making a distinguish between normal and abnormal samples. Also, they have interpreted the output of intermediate convolutional layers of a fully CNN to represent the whole video frames at one step. This method has effectively coped with challenges and the heavy overhead of patch based approaches.

\section{Deep Networks for AD}
\label{sec:sec3}
Joint learning of discriminative features, \ie, $p_\NN$ and $\DD$ all in one and as an end-to-end DNN, is one of the most prominent and essential characteristics where the previously discussed solutions conspicuously lack. Therefore, we now delve deeply into effective techniques: self-supervised learning, generative networks and anomaly generation. Note that learning a DNN to classify the normal and anomalous images or video frames is not straightforward. Unavailability of abnormal instances or unbalance training data imposes many difficulties to deal with such tasks. Hereupon, researchers have taken advantage of self-supervised learning and generative networks to learn an end-to-end deep network aiming to accurately detect the out-of-distribution data.

\subsection{Self-supervised Learning}
In real-world AD applications, the model merely accesses the normal data or minimal abnormalities. To train a deep end-to-end network on such data, researchers have tended to learn $p_\NN$ and $\DD$ implicitly. Accordingly, a neural network is trained under the specific constraints to learn $p_\NN$. Not satisfying the desired constraint(s) for a test sample, $X_{test}$, means that the sample does not follow $p_\NN$ and can be considered as an anomaly. Auxiliary tasks such as minimizing the Reconstruction Error (RE), forcing the latent representation to be sparse and predicting next frames of videos are well-known and popular self-supervised tasks to learn $p_\NN$ while $\DD$ is a score that shows how the specified constraint(s) are satisfied for detecting the outliers.

\noindent
\textbf{Encoder-Decoder Based Methods.}
The parameters of one or several deep encoder-decoder networks are learned to precisely reconstruct the training instances, \ie, normal data (${X}_\mathcal{N}$). These neural networks are trained by optimizing the Equation \ref{eq:re}.
\begin{equation}
    \mathbf{L}=\frac{1}{m}\sum{||{X}_\mathcal{N}^{i}-D(E({X}_\mathcal{N}^{i}))||^2}
\label{eq:re}
\end{equation}
Where $D(E(X))$ is an encoder-decoder network that implicitly learns the distribution of normal data, $p_\mathcal{N}$. The parameters of $D(E(X))$ is optimized to reconstruct normal events not abnormalities. Hence, reconstructing a sample with high RE, \ie, more than a predetermined threshold, shows that it is an irregularity \cite{tang2020integrating,xia2015learning,sabokrou2016video,zhai2016deep,sabokrou2019self,hasan2016learning,lu2013abnormal,luo2017revisit,nguyen2019anomaly,ravanbakhsh2017abnormal}. Directly using RE is the simplest idea for detecting out-of-distribution data by encoder-decoder networks. Although using just RE provide us with high performance in terms of accuracy, high false positive detection is problematic. To overcome this difficulty, several novelty detection methods have been introduced based on this idea. Most of them relied on manipulating or using the latent space of an auto-encoder to gain better results. \cite{park2020learning,gong2019memorizing} had a tendency to store several prototypes for normal samples by considering a memory module where individual items in it correspond to prototypical features of normal patterns. \cite{zhao2017spatio} used encode-decoder networks for modeling the spatio-temporal characteristics of videos. 

Although the RE is a useful criteria for distinguishing normality and abnormality, higher RE is not a conclusive proof for considering an instance as anomaly. An encoder-decoder networks can be used for the other purposes. It trained adversarially by \cite{sabokrou2018adversarially} for the pre-processing. Moreover, \cite{liu2018future} made use of an U-Net \cite{ronneberger2015u} as a deep encoder-decoder to predict the next frame of videos. In the training phase, both input (frame at time $t$ or $I_t$) and output (frame at time $t+1$ or $I_{t+1}$) of the neural network are normal, \ie, follow the distribution of normal frames ($p_{\mathcal{N}}$). In the test stage, a frame considered as an abnormal sample if it has a noticeable diversion from what was predicted. Encoder-decoder networks are also applicable to AD problem in medical imaging where \cite{baur2021autoencoders} carried out a comprehensive analysis of this area of research.

\noindent
\textbf{CNNs.}
Generally, due to the high dimensionality of images and video frames, learning two networks jointly, \ie, an encoder and a decoder, for mapping the input to a latent space and then recovering the original input is computationally expensive. To settle this issue, researchers decided to learn a CNN on available normal data for a pre-text task and detect the out-of-distribution data by analyzing the different responses of the neural networks to different types of input data (normal or anomaly). For example, \cite{golan2018deep} train a CNN as a classifier to recognize which geometric transformation is applied to the training data. In case of facing anomalous instances, the trained model is bewildered and output an in-confident result which can be considered as an abnormality. 

Obviously, minimizing the RE or predicting the next frame are not directly relevant to the AD problem. Nevertheless, leveraging the outcome of such tasks to track the behavior of input samples leading to effectively detecting anomalous events. In the aforementioned methods, $p_\NN$ is learned implicitly with the aim of optimizing the target task. $\DD$, that determine whether an instance is anomaly or not, is defined separately and not by the the neural network. Therefore, this network is not able to directly make a decision about the type of data. Consequently, taking this manner cannot be considered as an end-to-end setting.

\subsection{Generative Networks}
Although the above-mentioned techniques of deep networks for AD, including those are based on deep features and trained by self-supervised tasks, successfully derive a benefit from deep learning structures, neither of which is an end-to-end DNN. The absence of samples belonging to the outlier/abnormal class is the major obstacle for learning an end-to-end DNN. GANs is a very helpful tool to tackle this problem. They are broadly composed of two CNNs: Generator (G) and Discriminator (D). G tends to generate data instances with the same distribution of training data instances to fool D, in efforts to manipulate it into detecting G($X$) as real data, while D attempts to make a distinction between generated data by G and original training samples. These neural networks are jointly and adversarially trained with respect to the following objective function: 
\begin{equation}
\begin{aligned}
\min_G \max_D ~ & \Big( \mathbb{E}_{X \sim  P_\mathcal{N}}[\log(D(X))] \\
& + \mathbb{E}_{\tilde{X} \sim  P_\mathcal{N}+\mathcal{N}_\sigma}[\log(1-D(G(\tilde{X})))] \Big)
\end{aligned}
\label{eq:q2}
\end{equation}
In the training duration, G simultaneously generates irregularities for D network, and D is trained as a binary classifier. Eventually, D is capable of acting as an end-to-end one-class classifier. This is the basis of most of the end-to-end algorithms for AD tasks. In \cite{ravanbakhsh2017abnormal}, G is considered to be an encoder-decoder network exploited for reconstructing normal instances. The inability of G for reproducing the anomalous frames can be exploited for properly detecting the nature of input data. Even though this work take advantage of GANs, it still cannot be considered as an end-to-end method. In \cite{sabokrou2018adversarially}, not only the generator recovers (\ie, reconstructs) inliers, but it is also exploited for pre-processing to improve the performance of D as an end-to-end anomaly detector. GANs plays a significant role to effectively perform AD owing to the capability of learning an end-to-end network and generating abnormal samples at the same time. Thus, researchers has been encouraged to concentrate on GAN based methods \cite{kimura2020adversarial,sabokrou2018avid,sabokrou2018adversarially,schlegl2017unsupervised,zaheer2020old,ahmadi2019generative}. The GAN based AD solutions have achieved the notable performance, but on the other side of the coin, these approaches suffer from several issues making them inapplicable and ineffective in realistic problems. In a nutshell, expensive training, instability, difficulties in reproduction and mode collapse are the main downside of such methods. 

\noindent
\subsection{Anomaly Generation}
\label{sec:ag}
AD is a very critical and challenging task. On the other hand, the most prominent solutions, \ie, those of which exploiting GAN, is not reliable enough to be applied to real-world applications. Utilizing GANs for generating anomalous data, instead of directly using it, converts the problem of AD into a binary classification problem. This way of approaching the problem can also be utilized for abnormal data augmentation. This idea is presented by \cite{pourreza2021g2d}. The main contribution of this work is to train a Wasserstein GAN on normal instances and exploit its generator before the complete convergence. In this case, generated irregular data have a controlled deviation from normal samples. Generated abnormalities alongside the available normal data instances form an informative training set for the task of AD.

\section{Datasets}
\label{sec:dataset}
In this section, we briefly review the most widely-used and popular datastes have be used for the assessment process of image/video AD very .

\noindent
\textbf{Image.}
MNIST \cite{lecun2010mnist} is a large collection of 28$\times$28 gray-scale images of handwritten single digits between 0 and 9. Therefore, the total number of classes is 10 and all of the images are labeled.
CIFAR-10 and CIFAR-100 \cite{krizhevsky2009learning} consist of 10 and 100 classes of 32$\times$32 images, respectively. CIFAR-100 classes are non-overlapping and mutually exclusive with the CIFAR-10 classes.
Caltech-256 \cite{griffin2007caltech} includes 256 object categories and 30,607 images in total. Each category contains at least 80 images which is an improvement compared to Caltech-101 with the minimum number of 31 images per each group.
ImageNet \cite{deng2009imagenet} offers tens of millions of cleanly sorted images. It includes various subtrees in which there are thousands of synsets and millions of images.
MVTec \cite{bergmann2019mvtec} introduced unsupervised AD tasks in natural image data. It imitates the real-world scenarios and consists of 5,354 high-resolution images of five unique textures and ten unique objects from different domains.

\noindent
\textbf{Video.}
UMN\footnote{Available at http://mha.cs.umn.edu/} contains normal and anomalous events. normal events are about individuals wandering around and abnormality is characterized by only running action.
UCSD \cite{li2014ucsd}, is composed of two subsets: The UCSD Pedestrian 1 (Ped1) dataset and the UCSD Pedestrian 2 (Ped2) dataset. The resolutions of Ped1 and Ped2 are 158$\times$234 and 240$\times$360, respectively. Since the dominant mobile objects in these datasets are pedestrians, the rest objects such as cars, bicycles and skateboarders are considered as anomalies.
CUHK Avenue \cite{lu2013abnormal} includes 15,328 frames for training phase and 15,324 frames for the stage of testing with a resolution of 640$\times$360. It contains 47 different abnormalities, \eg, throwing objects and moving in unusual directions.
Train \cite{zaharescu2010anomalous} considers the behaviour of people in the train. The anomalous events are mainly correspond to strange movements of individuals in the train.
UCF-Crime \cite{sultani2018real} is a large-scale dataset consisting of 1,900 long and untrimmed manually collected real-world surveillance videos with 13 realistic anomalies such as explosion, stealing and accidents. The dataset can be utilized in two modes: general AD in which all of the data are grouped into normal and abnormal activities and specifically recognizing each of 13 anomalous activities. The former mode can be exploited as a binary classification task.
ShanghaiTech Campus \cite{luo2017revisit} is a very challenging dataset that includes 330 training videos and 107 testing ones with 130 anomalous events. The coverage of various types of anomalies is a distinguishing feature of this dataset which totally consists of 13 scenes.
Street Scene \cite{ramachandra2020street} is a dataset for AD problems that has noticeable labeled anomalous events and also different irregularity types for a single scene AD. This dataset contains a video of a two-way urban street including bike lanes and pedestrian sidewalks capturing a scene including large variety of activities with high resolution.

\section{Open Challenges and Future Directions}
Proposed methods using the new techniques have achieved superb performance compared to traditional solutions in both terms of complexity and accuracy. However, they encounter difficulties involving effort to devise a reliable approach that can be applied to real-world problems. Here, we have briefly pointed out the noteworthy facets of image/video AD which have been somehow neglected. 

\noindent
\textbf{Detection and False Positive Rate.} Ideally, an AD method should accurately detect outliers with a very low false positive rate. In reality, there are solutions with remarkable accuracy while they fail to effectively deal with high false alarm rate. The detection and false positive rate are directly related to the size of the selected region for processing. In other words, processing of pixels (small patches) and large patches results in high accuracy and high false positive rate, respectively. Presenting an approach with high detection rate, and at the same time, keeping the false positive rate low is a valuable research topic. Generally, some of normal samples or events occur in the training data with low frequency making them akin to abnormal samples. As a consequence, distinguishing such data from anomalies become a difficult task. To resolve this problem, Up-sampling the infrequent normal data helps to avoid the false detection.

\noindent
\textbf{Fairness.}
There is a bunch of reasons which are responsible for unfairness in AD tasks. Indeed, some issues such as skewed samples, limited features, tainted examples, disparity of sample size and proxies\cite{bolukbasi2016man} can bias the training set. This factor causes the output of a trained DNN not to be completely fair. \cite{zhang2020towards} has investigated that current DNNs like Deep Support Vector Data Description (DSVDD) \cite{ruff2018deep} has failed in fairness evaluation. Research on the fairness of AD methods and proposing a more efficient solution is an interesting research direction. 

\noindent
\textbf{Explanation of the AD Method.}
Researchers put effort into interpret DNNs with the aim of explaining their outcomes. In fact, not considering the DNN as a black-box approach leads to a better understandability and reliability \cite{holzinger2018machine}. Since AD is a very critical task, the presented works should be perfectly rational in the corresponding context. Furthermore, training the DNN should be human understandable. Working on solutions for explaining AD as a OCC problem is a step forward for the community. 

\noindent
\textbf{Object Interaction.}
video AD state-of-the-arts take advantage of different types of DNNs such as 3D-CNN and LSTM to learn both spatial and temporal features. These techniques can detect an anomalous object either has a different appearance or move differently in relation to other items. In videos, the appearance of objects should be real and logical as well as their interactions. Earlier methods are incapable of understanding the uncommon and unusual relationships between the objects in the scene. Hence, they are ineffective in properly detecting different kinds of anomalies.

\noindent
\textbf{Safety.}
DNNs are vulnerable to adversarial attacks. Only a targeted minor manipulation of input pixels confuses DNNs in a way that mistakenly classifies the input data \cite{goodfellow2014explaining}. For AD problems, adversarial attack does not mean detecting irregular events which in most cases, notably for surveillance cameras, causes severe damages. For supervised tasks, \eg, classification, improving the robustness of DNNs and defensing against such attacks are widely investigated while it does not receive enough attention for AD. \cite{adam2020robustness} has discussed adversarial attacks and also evaluate them merely on encoder-decoder based AD approaches. 

\noindent
\textbf{Adoptable AD.}
Regarding the previously mentioned concepts, the normal reference model, $p_\NN$, represents the common characteristics of regular events in a video. Samples with a enough deviation from such model detected as abnormalities. The concept of normal videos may gradually change due to different reasons. In such cases, $p_\NN$ no longer be valid for AD which is performed based on. To get over this hurdle, $p_\NN$ should gradually updated and incrementally learned. Continual Learning (CL) is an umbrella term for incremental learning and gradually updating the learned model. Considering the CL for AD tasks requires more attention by the researcher. Limited number of works are done on this topic \cite{doshi2020continual,stocco2020towards}. under realistic conditions, usually just one particular scene is required to deal with the task of AD. Hereupon, a novel few-shot scene-adaptive AD solution is presented by \cite{lu2020few}. The primary goal of this work is training a model to detect anomalous events in a previously unseen scene with only a few frames.

\noindent
\textbf{Generating Outliers.}
All previous methods for the OCC on images/videos have tended to learn a network on just normal data and formed a reference model for the target class instances. Learning a network to understand the distribution of outliers by only accessing to inliers and generating the meaningful samples rely on the learned distribution is an interesting procedure. In this way, by having access to both inlier and outlier data, the problem can be solved as a binary classification problem. As mentioned in previous section, several state-of-the-arts based on GANs are proposed for the OCC task. It is worth mentioning that, the discriminator of such network can be used as a irregularity detector. Nevertheless, such solutions poses difficulties for the training procedure and needs trail and error. We struggle with this issue in our previous works such as AVID \cite{sabokrou2018avid} and ALOCC \cite{sabokrou2018adversarially}. Albeit, generating abnormal data is not a straightforward process. \cite{pourreza2021g2d} only generates very simple images. It has exploited the inability of the generator of the GAN for producing anomalies. In this way, most of the generated data are very similar to the random noises. For generating more complex images and also abnormal video frames, a more efficient approach should be investigated. 

\noindent
\textbf{Realistic Datasets.}
The popular and standard datasets for the AD tasks in visual data are broadly captured under controlled conditions which are far from the reality. For example, UMN dataset is very simple where the performance of methods is saturated on it. Furthermore, videos in UCSD Ped1 and Ped2 datasets are captured in just one location and thus the camera is fixed during the training procedure. The resolution of video frames is extremely low. In fact, anomalies in those videos are considered very simple so that the real-world anomalous events are not properly reflected in video surveillance. Recently, a more real dataset is introduced by \cite{sultani2018real}. 

\noindent
\textbf{Early Detection or Prediction.}
Compensation for late detection of abnormal events in videos is costly. Most of the previous works attempt to detect video anomalies very accurately. These approaches are able to detect irregular activities in videos which are either over or near to end. With respect to the nature of this task, early detection of such events is very critical. A well-timed alarm can prevent or at least reduce the loss or damages caused by occurring anomalous activities. Besides, predicting irregular activities is an exciting task that AD systems can derive a benefit from. For example, if a surveillance camera is capable of predicting an imminent accident on a highway, a traumatic event can be prevented by broadcasting a timely alarm. Though there are many research works on the activity recognition, abnormal activity recognition is not well studied. To the best of our knowledge, there is not a standard benchmark for anomaly prediction in videos. Accordingly, it is very difficult to investigate this controversial issue.

\vspace{-2mm}
\section{Conclusions}
The absence of a specialized review of deep learning based AD for images and videos motivated us to present a comprehensive survey in this research area. The main concentration of our work is on unsupervised techniques. we provide a precise definition of AD concept along with a thorough categorization of recent proposed methods for AD. Afterward, the open challenges are carefully explained in order to investigate formidable obstacles and useful directions for future works.

\bibliographystyle{named}
\bibliography{References}

\end{document}